\documentclass[a4paper,11pt,twoside]{article}
\usepackage{mic09}
\usepackage{graphicx}

\usepackage{url}
\usepackage{ams} 
\usepackage{algorithmic} 
\usepackage{algorithm} 

\def\id{195} 

\begin{document}

\mictitle{Improvements for multi-objective flow shop scheduling by
{P}areto {I}terated {L}ocal {S}earch}

\author{
  \micauthor{Martin Josef}{Geiger}{\firstaffiliationmark}
}

\institutions{
  \micinstitution{\firstaffiliationmark}
  {Logistics Management Department,\\Helmut Schmidt University --- University of the Federal Armed Forces Hamburg}
  {Holstenhofweg 85, 22043 Hamburg, Germany}
  {mjgeiger@hsu-hh.de}
}

\maketitle
\thispagestyle{fancyplain}

\section{\label{sec:mo:scheduling}Introduction}%

The flow shop scheduling problem consists in the assignment of a set
of jobs $\mathcal{J} = \{ J_{1}, \ldots, J_{n} \}$, each of which
consists of a set of operations $J_{j} = \{ O_{j1}, \ldots,
O_{jo_{j}} \}$ onto a set of machines $\mathcal{M} = \{ M_{1},
\ldots, M_{m} \}$ \cite{blazewicz:2001:book,pinedo:2002:book}. Each
operation $O_{jk}$ is processed by at most one machine at a time,
involving a non-negative processing time $p_{jk}$. The result of the
problem resolution is a schedule $x$, defining for each operation
$O_{jk}$ a starting time $s_{jk}$ on the corresponding machine.
Several side constraints are present which have to be respected by
any solution $x$ belonging to the set of feasible schedules $X$.
Precedence constraints $O_{jk} \rhd O_{jk+1} \forall j = 1, \ldots,
n, k = 1, \ldots, o_{j}-1$ between the operations of a job $J_{j}$
assure that processing of $O_{jk+1}$ only commences after completion
of $O_{jk}$, thus $s_{jk+1} \geq s_{jk} + p_{jk}$. In flow shop
scheduling, the machine sequence in which the operations are
processed by the machines is identical for all jobs, and for the
specific case of the permutation flow shop scheduling the job
sequence must also be the same on all machines.

The assignment of starting times to the operations has to be done
with respect to one or several optimality criteria. Most optimality
criteria are functions of the completion times $C_{j}$ of the jobs
$J_{j}$, $C_{j} = s_{jo_{j}} + p_{jo_{j}}$

The most prominent optimality criteria are the maximum completion
time (makespan) $C_{max} = \max C_{j}$ and the sum of the completion
times $C_{sum} = \sum_{j=1}^{n} C_{j}$. Others express violations of
due dates $d_{j}$ of jobs $J_{j}$. A due date $d_{j}$ defines a
latest point of time until production of a job $J_{j}$ should be
completed as the finished product has to be delivered to the
customer on or up to this date. A possible optimality criteria based
on tardiness of jobs is e.\,g.\ the total tardiness $T_{sum} =
\sum_{j=1}^{n} T_{j}$, where $T_{j} =  \max (C_{j} - d_{j}, 0)$.

It is known, that for \emph{regular} optimality criteria at least
one \emph{active} schedule $x$ does exist which is also optimal. The
representation of an active schedule is possible using a permutation
of jobs $\pi = ( \pi_{1}, \ldots, \pi_{n} )$, where each $\pi_{j}$
stores a job $J_{k}$ at position $j$. The permutation is then
decoded into an active schedule by assuming the starting times of
all operations as early as possible with respect to the precedence
constraints and the given sequence in $\pi$. As a consequence, the
search is, instead of searching all possible schedules, restricted
to the much smaller set of active schedules only.

Multi-objective approaches to scheduling consider a vector $G(x) =
(g_{1}(x), \ldots, g_{K}(x))$ of optimality criteria at once
\cite{tkindt:2002:book}. As the relevant optimality criteria are
often of conflicting nature, not a single solution $x \in X$ exists
optimizing all components of $G(x)$. Optimality in multi-objective
optimization problems is therefore understood in the sense of
Pareto-optimality, and the resolution of multi-objective
optimization problems lies in the identification of all elements
belonging to the Pareto set $P$, containing all alternatives $x$
which are not dominated by any other alternative $x' \in X$. 


Several approaches of metaheuristics have been formulated and tested
in order to solve the permutation flow shop scheduling problem under
multiple, in most cases two, objectives. Common to all is the
representation of solutions using permutations $\pi$ of jobs, as in
previous investigations only regular functions are considered.\\
First results have been obtained using Evolutionary Algorithms,
which in general play a dominant role in the resolution of
multi-objective optimization problems when using metaheuristics.
This is mainly due to the fact that these methods incorporate the
idea of a set of solutions, a so called \emph{population}, as a
general ingredient. Flow shop scheduling problems minimizing the
maximum completion time and the average flow time have been solved
by {\sc Nagar}, {\sc Heragu} and {\sc Haddock}
\cite{nagar:1996:article}. In their work, they however combine the
two objectives into a weighted sum. Problems minimizing the maximum
completion time and the total tardiness are solved by {\sc Murata},
{\sc Ishibuchi} and {\sc Tanaka} \cite{murata:1996:article}, again
under the combination of both objectives into a weighted sum. Later
work on the same problem class by {\sc Basseur}, {\sc Seynhaeve} and
{\sc Talbi} \cite{basseur:2002:inproceedings} avoids the weighted
sum approach, using dominance relations among the solutions
only.\\Most recent work is presented by {\sc Loukil}, {\sc Teghem}
and {\sc Tuyttens} \cite{loukil:2005:article}. Contrary to
approaches from Evolutionary Computations, the authors apply the
Multi Objective Simulated Annealing approach MOSA of {\sc Ulungu},
{\sc Teghem}, {\sc Fortemps} and {\sc Tuyttens}
\cite{ulungu:1999:article} to a variety of bi-criterion scheduling
problems.\\
Flow shop scheduling problems with three objectives are studied by
{\sc Ishibuchi} and {\sc Murata} \cite{ishibuchi:1998:article}, and
{\sc Ishibuchi}, {\sc Yoshida} and {\sc Murata}
\cite{ishibuchi:2003:article}. The authors minimize the maximum
completion time, the total completion time, and the maximum
tardiness at once. A similar problem minimizing the maximum
completion time, the average flow time, and the average tardiness is
then tackled by {\sc Bagchi}
\cite{bagchi:1999:book,bagchi:2001:inproceedings}.

\section{\label{sec:pils}Pareto Iterated Local Search} %
The Pareto Iterated Local Search (PILS) metaheuristic is a concept
for the solution of multi-objective optimization problems. It
combines the two main driving forces of local search,
intensification and diversification, into a single algorithm, and
extends the work presented in \cite{geiger:2007:ejor}. The
motivation behind this concept can be seen in the increasing demand
for simple, yet effective heuristics for the resolution of complex
multi-objective optimization problems. Two developments in local
search demonstrate the effectiveness of some intelligent ideas that
make use of certain structures within the search space topology of
problems:

\begin{enumerate}
\item Iterated Local Search, introducing the
idea of perturbing solutions to overcome local optimality and
continue search in interesting areas of the search space
\cite{lourenco:2003:incollection}. After the pioneering work of {\sc
Boese} \cite{boese:1996:phdthesis}, who investigated properties of
the search space of the traveling salesman problem, this concept has
been used with increasing success on problems where solutions of
high quality can be found relatively concentrated in alternative
space.
\item Second, Variable Neighborhood Search
\cite{hansen:2003:incollection}, combining multiple neighborhood
operators into a single algorithm in order to avoid local optimality
in the first place.
\end{enumerate}

In the proposed concept, both paradigms are combined and extended
within a search framework handling not only a single but a set of
alternatives at once.

\begin{figure}[!ht]
\begin{center}
\includegraphics{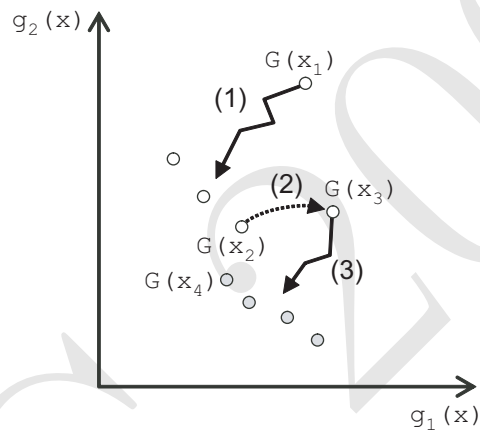}
\end{center}
\caption{\label{fig:pils}Illustration of the Pareto Iterated Local
Search metaheuristic. The archive of the currently best solutions is
updated during the search. Here, $G(x_{4})$ dominates $G(x_{2})$
which is going to be deleted from $P^{approx}$.}
\end{figure}

The main principle of the algorithm is sketched in Figure
\ref{fig:pils}. Starting from an initial solution $x_{1}$, an
improving, intensifying search is performed until a set of locally
optimal alternatives is identified, stored in a set $P^{approx}$
representing the approximation of the true Pareto set $P$. No
further improvements are possible from this point. In this initial
step, a set of neighborhoods ensures that all identified
alternatives are locally optimal not only to a single but to a set
of neighborhoods. This principle, known from Variable Neighborhood
Search, promises to lead to better results as it is known that all
global optima are also locally optimal with respect to all possible
neighborhoods while this is not necessarily the case for local
optima.

After the identification of a locally optimal set, a diversification
step is performed on a solution $x_{2}$ using a perturbation
operator, continuing search from the perturbed solution $x_{3}$. The
perturbation operator has to be significantly different from the
neighborhoods used in intensification, as otherwise the following
search would return to the previous solution. On the other hand
however, the perturbation should not entirely destroy the
characteristics of the alternative. Doing that would result in a
random restart of the search without keeping promising attributes of
solutions.

The PILS metaheuristic may be formalized as given in
Algorithm~\ref{alg:pils}. The intensification of the algorithm,
illustrated in steps (1) and (3) of Figure~\ref{fig:pils} is between
the lines 6 to 21, the description of the diversification, given in
step (2) of Figure~\ref{fig:pils} is between the lines 22 to 26.

\begin{algorithm}[!ht]
\caption{\label{alg:pils}Pareto Iterated Local Search}
\begin{algorithmic}[1]
\STATE{Initialize control parameters: Define the neighborhoods $\mathbf{N}_{1}, \ldots, \mathbf{N}_{k}$}%
\STATE{Set $i = 1$} %
\STATE{Generate initial solution $x$}%
\STATE{Set $P^{approx} = \{ x \}$}%
\REPEAT%
    \REPEAT%
        \STATE{Compute $\mathbf{N}_{i}(x)$}%
        \STATE{Evaluate $\mathbf{N}_{i}(x)$}%
        \STATE{Update $P^{approx}$ with $\mathbf{N}_{i}(x)$}%
        \IF{$\exists x' \in \mathbf{N}_{i}(x) \mid x' \preceq x$}%
            \STATE{Set $x = x'$}%
            \STATE{Set $i = 1$}%
            \STATE{Rearrange the neighborhoods $\mathbf{N}_{1}, \ldots, \mathbf{N}_{k}$ in some random order}%
        \ELSE%
            \STATE{Set $i = i + 1$}%
        \ENDIF%
    \UNTIL{$x$ locally optimal with respect to $\mathbf{N}_{1}, \ldots, \mathbf{N}_{k}$, therefore $i > k$}%
    \STATE{Set neighborhoods of $x$ as \lq{}investigated\rq{}}
    \STATE{Set $i = 1$}
    \IF{$\exists x' \in P^{approx} \mid$ neighborhoods not investigated yet}%
        \STATE{Set $x = x'$}%
    \ELSE%
        \STATE{Randomly select some $x' \in P^{approx}$}%
        \STATE{Compute $x'' = \mathbf{N}_{perturb}(x')$}%
        \STATE{Set $x = x''$}%
    \ENDIF%
\UNTIL{termination criterion is met}%
\end{algorithmic}
\end{algorithm}

It can be seen, that the algorithm computes a set of neighborhoods
for each alternative. The sequence in which the neighborhoods are
computed is arranged in a random fashion, described in line 13 of
Algorithm~\ref{alg:pils}. This introduces an additional element of
diversity to the concept, as otherwise the search might be biased by
a certain sequence of neighborhoods.

\section{\label{sec:application}Experiments on multi-objective flow shop scheduling problems} %
\subsection{Algorithm configuration and experimental setup} %
In the following, the Pareto Iterated Local Search is applied to a
set of benchmark instances of the multi-objective permutation flow
shop scheduling problem. The first instances have been provided by
{\sc Basseur}, {\sc Seynhaeve} and {\sc Talbi}
\cite{basseur:2002:inproceedings}, who defined due dates for the
well-known instances of {\sc Taillard} \cite{taillard:1993:article}.
The instances range from $n = 20$ jobs that have to be processed on
$m = 5$ machines to $n = 100, m = 20$. All of them are solved under
the simultaneous consideration of the minimization of the maximum
completion time $C_{max}$ and the total tardiness $T_{sum}$ and are
referred to as \lq{}Ta $n \times m$\rq{}.\\We also solved the
benchmark instance \lq{}Ba $49 \times 15$\rq{} by {\sc Bagchi}
\cite{bagchi:1999:book}, consisting of $n = 49$ jobs on $m = 15$
machines. The three objective functions of the data set are the
minimization of the maximum completion time $C_{max}$, the
minimization of the average completion time $\frac{1}{n}C_{sum}$,
and the minimization of the average tardiness $\frac{1}{n}T_{sum}$.

Three operators are used in the definition of the neighborhoods
$\mathbf{N}_{1}, \ldots, \mathbf{N}_{k}$, described in the work of
{\sc Reeves} \cite{reeves:1999:article}. First, an exchange
neighborhood, exchanging the position of two jobs in $\pi$, second,
a forward shift neighborhood, taking a job from position $i$ and
reinserting it at position $j$ with $j < i$, and finally a backward
shift neighborhood, shifting a job from position $i$ to $j$ with $j
< i$. All operators are problem independent operators for
permutation-based representations, each computing $\frac{n(n-1)}{2}$
neighboring solutions.

After a first approximation $P^{approx}$ of the Pareto set is
obtained, one element $x' \ P^{approx}$ is selected by random and
perturbed into another solution $x''$. We use a special neighborhood
that on one hand leaves most of the characteristics of the perturbed
alternatives intact, while still changes the positions of some jobs.
Also, several consecutive applications of the neighborhoods
$\mathbf{N}_{1}, \ldots, \mathbf{N}_{k}$ would be needed to
transform $x''$ back into $x'$. This is important, as otherwise the
algorithm might return to the initially perturbed alternative $x'$,
leading to a cycle in the search path.\\The perturbation
neighborhood $\mathbf{N}_{perturb}$ can be described as follows.
First, a subset of $\pi$ is randomly selected, comprising four
consecutive jobs at positions $j, j+1, j+2, j+3$. Then a neighboring
solution $x''$ is generated by moving the job at position $j$ to
$j+3$, the one at position $j+1$ to $j+2$, the one at position $j+2$
to $j$, and the job at position $j+3$ to $j+1$, leaving the jobs at
the positions $<j$ and $>j+3$ untouched. In brief, this leads to a
combination of several exchange and shift moves, executed at once.


In order to analyze the quality of the approximations, we compare
the results obtained by PILS to the approximations of a
multi-objective multi-operator search algorithm MOS, described in
Algorithm \ref{alg:mos}.

\begin{algorithm}[!ht]
\caption{\label{alg:mos}Multi-objective multi-operator search
framework}
\begin{algorithmic}[1]
\STATE{Generate initial solution $x$, set $P^{approx} = \{ x \}$}%
\REPEAT%
    \STATE{Randomly select some $x \in P^{approx} \mid $ neighborhoods not investigated yet}%
    \STATE{Randomly select some neighborhood $\mathbf{N}_{i}$ from $\mathbf{N}_{1}, \ldots, \mathbf{N}_{k}$}%
    \STATE{Generate $\mathbf{N}_{i}(x)$}%
    \STATE{Update $P^{approx}$ with $\mathbf{N}_{i}(x)$}%
    \IF{$x \in P^{approx}$}%
        \STATE{Set neighborhoods of $x$ as \lq{}investigated\rq{}}%
    \ENDIF%
\UNTIL{$\not\!\exists x \in P^{approx} \mid$ neighborhoods not investigated yet}%
\STATE{Return $P^{approx}$}
\end{algorithmic}
\end{algorithm}

The MOS Algorithm,
taken from \cite{geiger:2007:ejor}, 
is based on the concept of Variable Neighborhood Search, extending
the general idea of several neighborhood operators by adding an
archive $P^{approx}$ towards the optimization of multi-objective
problems. For a fair comparison, the same neighborhood operators are
used as in the PILS algorithm. After the termination criterion is
met in step 10, we restart search while keeping the approximation
$P^{approx}$ for the final analysis of the quality of the obtained
solutions.

\subsection{Results} %
The benchmark instances of {\sc Basseur} and the one of {\sc Bagchi}
have been solved using the PILS algorithm. In each of the 100 test
runs, the approximation quality of the obtained results has been
analyzed using the $D_{1}$ and $D_{2}$ metrics of {\sc Czy\.{z}ak}
and {\sc Jaszkiewicz} \cite{czyzak:1998:article}. The two metrics
have been chosen for the analysis as they provide an interesting
interpretation from an economical point of view. Based on a so
called \lq{}achievement scalarizing function\rq{}, they compute the
average ($D_{1}$) and the maximum ($D_{2}$) regret a decision maker
would have to face when trying to select a certain most preferred
alternative $x^{*} \in P$, approximated by the results in
$P^{approx}$.

While for the smaller instances the optimal solutions are known, the
analysis for the larger instances has to rely on the best known
results published in the literature. Experiments have been carried
out on a Intel Pentium IV processor, running at 1.8 GHz.
Table~\ref{tbl:termination:criterion} gives an overview about the
number of evaluations executed for each instance. Clearly,
considerable more alternatives have to be evaluated with increasing
size of the problem instances to allow a convergence of the
algorithm. Also, the running times, given in
Table~\ref{tbl:termination:criterion}, too, increase with increasing
size of the problem instances. 
No significant difference in running behavior can be found when
comparing the two metaheuristics PILS and MOS. Apart from some minor
differences around the perturbation neighborhood
$\mathbf{N}_{perturb}$, the approaches are identical with respect to
the impact on the resulting running times as they use the same
neighborhood operators.

\begin{table}[!ht]
\begin{center}
\caption{\label{tbl:termination:criterion}Number of evaluations and
running times for the investigated instances.} 

\begin{tabular}{lrrr}\\
\hline %
Instance $n \times m$  & No of evaluations & Eval. time & Neighbor.\\
 & & & comp. time\\
\hline %
Ta $20 \times 5$ (\#1)  & 1,000,000 &   42.7 & 0.5\\
Ta $20 \times 5$ (\#2)  & 1,000,000 &   42.5 & 0.6\\
Ta $20 \times 10$ (\#1) & 1,000,000 &   78.0 & 0.6\\
Ta $20 \times 10$ (\#2) & 1,000,000 &   83.3 & 0.6\\
Ta $20 \times 20$     & 1,000,000   &  143.5 & 0.6\\
Ta $50 \times 5$      & 5,000,000   &  195.1 & 1.3\\
Ta $50 \times 10$     & 5,000,000   &  386.1 & 1.3\\
Ta $50 \times 20$     & 5,000,000   &  754.3 & 1.3\\
Ta $100 \times 10$    & 10,000,000  & 1459.7 & 2.5\\
Ta $100 \times 20$    & 10,000,000  & 2885.3 & 2.5\\
Ba $49 \times 15$     & 5,000,000   &  566.3 & 1.2\\
\hline%
\multicolumn{4}{l}{Times are given in milliseconds.}
\end{tabular}

\end{center}
\end{table}

An implementation of the algorithm has been made available within
an integrated software 
for the resolution of multi-objective scheduling problems using
metaheuristics. The system is equipped with an extensive user
interface that allows an interaction with a decision maker 
and is able to visualize the obtained results in alternative and
outcome space. The system also allows the comparison of results
obtained by different metaheuristics.

The average values obtained by the investigated metaheuristics are
given in Table \ref{tbl:results}. It can be seen, that PILS leads
for all investigated problem instances to better results for both
the D$_{1}$ and the $D_{2}$ metric. This general result is
consistently independent from the actual problem instance and
significant at a level of significance of $0.01$. For a single
instance, the \lq{}Ta $20 \times 5$~(\#1)\rq{}, PILS was able to
identify all optimal solutions in all test runs, leading to average
values of $D_{1} = D_{2} = 0.0000$. Apparently, this instance is
comparably easy to solve.

\begin{table}[!ht]
\begin{center}
\caption{\label{tbl:results}Average results of $D_{1}$ and
$D_{2}$}
\begin{tabular}{lrrrr}\\
\hline%
& \multicolumn{2}{c}{$D_{1}$} & \multicolumn{2}{c}{$D_{2}$}\\
Instance $n \times m$ & PILS & MOS & PILS & MOS\\
\hline
Ta $20 \times 5$ (\#1)  & 0.0000 & 0.0323 & 0.0000 & 0.1258\\
Ta $20 \times 5$ (\#2)  & 0.1106 & 0.1372 & 0.3667 & 0.4249\\
Ta $20 \times 10$ (\#1) & 0.0016 & 0.0199 & 0.0146 & 0.0598\\
Ta $20 \times 10$ (\#2) & 0.0011 & 0.0254 & 0.0145 & 0.1078\\
Ta $20 \times 20$     & 0.0088 & 0.0286 & 0.0400 & 0.1215\\
Ta $50 \times 5$      & 0.0069 & 0.0622 & 0.0204 & 0.1119\\
Ta $50 \times 10$     & 0.0227 & 0.3171 & 0.0897 & 0.4658\\
Ta $50 \times 20$     & 0.0191 & 0.3966 & 0.0616 & 0.5609\\
Ta $100 \times 10$    & 0.0698 & 0.3190 & 0.1546 & 0.4183\\
Ta $100 \times 20$    & 0.0013 & 0.2349 & 0.0255 & 0.3814\\
Ba $49 \times 15$     & 0.0202 & 0.2440 & 0.0701 & 0.3414\\
\hline
\end{tabular}

\end{center}
\end{table}

While it was possible to show in \cite{geiger:2007:ejor} that the
MOS algorithm is competitive to different Evolutionary Algorithms,
iterating search in qualitatively good areas of the search space by
PILS improves the results even further. Recalling, that with
increasing problem size an increasing amount of time is needed to
evaluate the alternatives, iterating in promising regions becomes
even more interesting as opposed to restarting the search.

\begin{figure}[!ht]
\begin{center}
\begin{tabular}{cc}
\includegraphics[width=8cm]{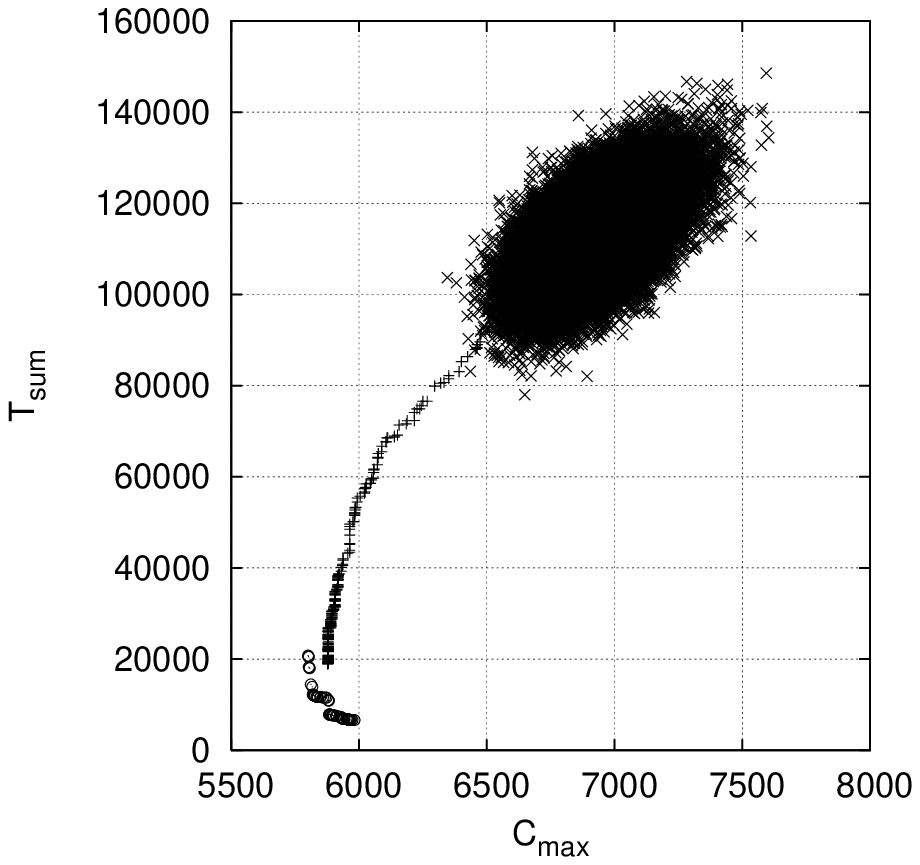} &
\includegraphics[width=8cm]{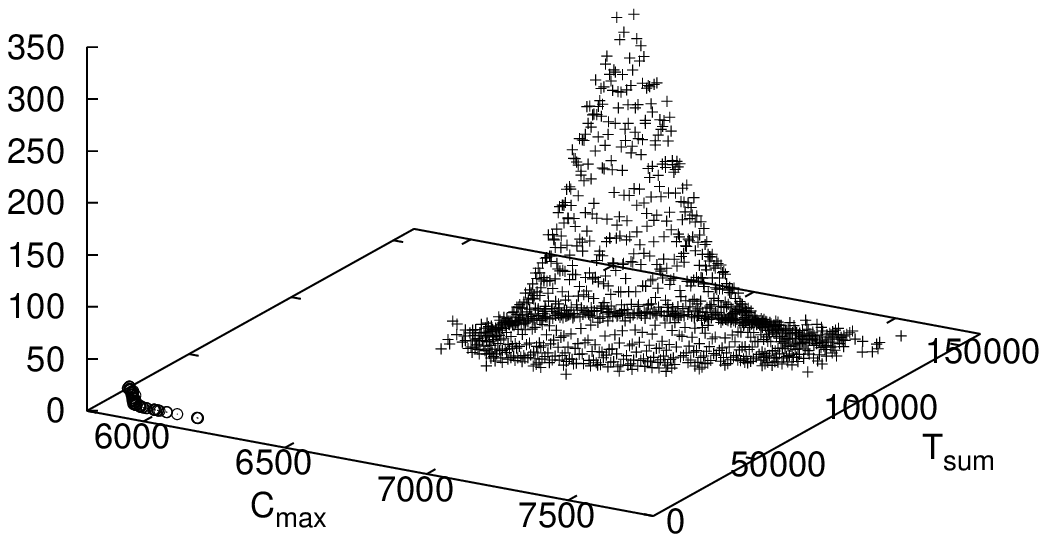}\\
(a) & (b)
\end{tabular}
\end{center}
\caption{\label{fig:AB} (a) Randomly generated solutions ($\times$),
intensification of search($+$), and Pareto front ($\odot$). (b)
Distribution of randomly generated solutions ($+$) compared to the
Pareto front ($\odot$)}
\end{figure}

A deeper analysis has been performed to monitor the resolution
behavior of the local search algorithms and to get a better
understanding of how the algorithm converges towards the Pareto
front. Figure \ref{fig:AB} (a) plots with symbol $\times$ the
results obtained by random sampling 50,000 alternatives for the
problem instance \lq{}Ta $100 \times 10$\rq{}, and compares the
points obtained during the first intensification procedure of PILS
until a locally optimal set is identified. The alternatives computed
starting from a random initial solution towards the Pareto front are
plotted as $+$, the Pareto front as $\odot$. It can be seen, that in
comparison to the initial solution even a simple local search
approach converges in rather close proximity to the Pareto front.
With increasing number of computations however, the steps towards
the optimal solutions get increasingly smaller, as it can be seen
when monitoring the distances between the $+$ symbols. After
convergence towards a locally optimal set, overcoming local
optimality is then provided by means of the perturbation
neighborhood $\mathbf{N}_{perturb}$.

An interesting picture is obtained when analyzing the distribution
of the randomly sampled 50,000 alternatives for instance \lq{}Ta
$100 \times 10$\rq{}. In Figure~\ref{fig:AB} (b), the number of
alternatives with a certain combination of objective function values
are plotted and compared to the Pareto front, given in the left
corner. It turns out that many alternatives are concentrated around
some value combination in the area of approximately $C_{max} =
6900$, $T_{sum} = 111500$, relatively far away from the Pareto
front.

When analyzing the convergence of local search heuristics toward the
globally Pareto front as well as towards locally optimal
alternatives, the question arises how many local search steps are
necessary until a locally optimal alternative is identified. From a
different point of view, this problem is discussed in the context of
computational complexity of local search
\cite{johnson:1988:article}. It might be worth investigating this
behavior in quantitative terms. Table
\ref{tbl:anz:bis:lokal:optimal} gives the average number of
evaluations that have been necessary to reach a locally optimal
alternative from some randomly generated initial solution. The
analysis reveals that the computational effort grows exponentially
with the number of jobs $n$.

\begin{table}[!ht]
\begin{center}
\caption{\label{tbl:anz:bis:lokal:optimal}Average number of
evaluations until a locally optimal alternative is reached}

\begin{tabular}{lrr|lrr}\\
\hline %
Instance $n \times m$   & No of jobs & No of evaluations & Instance $n \times m$   & No of jobs & No of eval.\\
\hline %
Ta $20 \times 5$ (\#1)  & 20  & 3,614 & Ta $50 \times 5$      & 50  & 53,645\\
Ta $20 \times 5$ (\#2)  & 20  & 3,292 & Ta $50 \times 10$     & 50  & 55,647\\
Ta $20 \times 10$ (\#1) & 20  & 2,548 & Ta $50 \times 20$     & 50  & 38,391\\
Ta $20 \times 10$ (\#2) & 20  & 2,467 & Ta $100 \times 10$    & 100 & 793,968\\
Ta $20 \times 20$       & 20  & 2,657 & Ta $100 \times 20$    & 100 & 479,420\\
& & & Ba $49 \times 15$     & 49  & 28,908\\
\hline%
\end{tabular}

\end{center}
\end{table}

\section{\label{sec:conclusions}Conclusions} 
In the past years, considerable progress has been made in solving
complex multi-objective optimization problems. Effective
metaheuristics have been developed, providing the possibility of
computing approximations to problems with numerous objectives and
complex side constraints. While many approaches are of increasingly
effectiveness, complex parameter settings are however required to
tune the solution approach to the given problem at hand.

The algorithm presented in this paper proposed a metaheuristic,
combining two recent principles of local search, Variable
Neighborhood Search and Iterated Local Search. The main motivation
behind the concept is the easy yet effective resolution of
multi-objective optimization problems with an approach using only
few parameters.

After an initial introduction to the problem domain of flow shop
scheduling under multiple objectives, the introduced PILS algorithm
has been applied to a set of scheduling benchmark instances taken
from literature. We have been able to obtain encouraging results,
despite the simplicity of the algorithmic approach. A comparison of
the approximations of the Pareto sets has been given with a
multi-operator local search approach, and, as a conclusion, PILS was
able to lead to consistently better results. We had however to
observe, that with increasing problem size, the number of iterations
needed to identify an only locally optimal solutions grows
exponentially.\\Nevertheless, the presented approach seems to be a
promising tool for the effective resolution of multi-objective
optimization problems. After first tests on problems from the domain
of scheduling, the resolution behavior on problems from other areas
might be an interesting direction for further research.

\bibliography{lit_bank,lit_bank_datei,lit_bank_nv,MJGeigerBIB}
\bibliographystyle{plain}

\end{document}